\documentclass{article}

     \PassOptionsToPackage{numbers, compress}{natbib}

    \usepackage[preprint]{neurips_2019}

\usepackage[utf8]{inputenc} 
\usepackage[T1]{fontenc}    
\usepackage{hyperref}       
\usepackage{url}            
\usepackage{booktabs}       
\usepackage{amsfonts}       
\usepackage{nicefrac}       
\usepackage{microtype}      
\usepackage{graphicx}
\usepackage{amsthm}
\usepackage{mathtools}
\usepackage{algorithm,algorithmic}
\usepackage{amssymb, multirow, paralist, color}
\newtheorem{thm}{Theorem}

\newtheorem{lem}{Lemma}

\newtheorem{defn}{Definition}
\newtheorem{ass}{Assumption}
\usepackage{enumitem}
\usepackage{booktabs}
\usepackage{amsmath}
\usepackage{bbm}

\title{A Distributed Training Algorithm of Generative Adversarial Networks with Quantized Gradients}

\author{%
  Xiaojun Chen$^{\dagger}$, Shu Yang$^{\dagger}$, Li Shen$^{\ddagger}$, Xuanrong Pang$^{\dagger}$\\
  $^\dagger$College of Computer Science and Software Engineering, Shenzhen University, Shenzhen, P.R.China\\
  $^\ddagger$Tencent AI Lab, Shenzhen, P.R.China\\
  \texttt{xjchen@szu.edu.cn, yangshu2018@email.szu.edu.cn} \\
  \texttt{mathshenli@gmail.com, pangxuanrong2018@email.szu.edu.cn}
}

\begin{document}

\maketitle

\begin{abstract}
Training generative adversarial networks (GAN) in a distributed fashion is a promising technology since it is contributed to training GAN on a massive of data efficiently in real-world applications. However, GAN is known to be difficult to train by SGD-type methods (may fail to converge) and the distributed SGD-type methods may also suffer from massive amount of communication cost. In this paper, we propose a {distributed GANs training algorithm with quantized gradient, dubbed DQGAN,} which is the first distributed training method with quantized gradient for GANs. The new method trains GANs based on a specific single machine algorithm called Optimistic Mirror Descent (OMD) algorithm, and is applicable to any gradient compression method that satisfies a general $\delta$-approximate compressor. The error-feedback operation we designed is used to compensate for the bias caused by the compression, and moreover, ensure the convergence of the new method. Theoretically, we establish the non-asymptotic convergence of {DQGAN} algorithm to first-order stationary point, which shows that the proposed algorithm can achieve a linear speedup in the parameter server model. Empirically, our experiments show that our {DQGAN} algorithm can reduce the communication cost and save the training time with slight performance degradation on both synthetic and real datasets.
\end{abstract}

\section{Introduction}
Generative adversarial networks (GANs) ~\cite{goodfellow2014generative}, a kind of deep generative model, are designed to approximate the probability distribution of a massive of data. It has been demonstrated powerful for various tasks including images generation~\cite{radford2016unsupervised}, video generation~\cite{vondrick2016generating}, and natural language processing~\cite{lin2017adversarial}, due to its ability to handle complex density functions. GANs are known to be notoriously difficult to train since they are non-convex non-concave min-max problems. Gradient descent-type algorithms are usually used for training generative adversarial networks, such as Adagrad~\cite{duchi2011adaptive}, Adadelta~\cite{zeiler2012adadelta}, RMSprop~\cite{tieleman2012lecture} and Adam~\cite{kingma2014adam}. However, it recently has been demonstrated that gradient descent type algorithms are mainly designed for minimization problems and may fail to converge when dealing with min-max problems~\cite{mertikopoulos2019optimistic}. 
In addition, the learning rates of gradient type algorithms are hard to tune when they are applied to train GANs due to the absence of universal evaluation criteria like loss increasing or decreasing. Another critical problem for training GANs is that they usually contain a large number of parameters, thus may cost days or even weeks to train. 

During the past years, much effort has been devoted for solving the above problems. Some recent works show that the Optimistic Mirror Descent (OMD) algorithm has superior performance for the min-max problems~\cite{mertikopoulos2019optimistic, daskalakis2018training, gidel2019a}, which introduces an extra-gradient step to overcome the the divergence issue of gradient descent-type algorithms for solving the min-max problems. Recently, Daskalakis et al.~\cite{daskalakis2018training} used OMD for training GAN in single machine. In addition, a few work also develop the distributed extragradient descent type algorithms to accelerate the training process of GANs \cite{jin2016scale, lian2017can, shen2018towards, tang2018d}. Recently, Liu et al.~\cite{liu2019decentralized} have proposed distributed extragradient methods to train GANs with the decentralized network topology, but their method suffers from the large amount of parameters exchange problem since they did not compress the transmitted gradients.

In this paper, we propose a novel distributed training method for GANs, named as the 
{Distributed Quantized Generative Adversarial Networks (DQGAN)}. The new method is able to accelerate the training process of GAN in distributed environment and a quantization technique is used to reduce the communication cost. To the best of our knowledge, this is the first distributed training method for GANs with quantization. The main contributions of this paper are summarized as three-fold:
\begin{itemize}
    \item We have proposed a {distributed optimization algorithm for training GANs}. Our main novelty is that the gradient of communication is quantified, so that our algorithm has a smaller communication overhead to further improve the training speed. Through the error compensation operation we designed, we can solve the convergence problem caused by the quantization gradient.
    \item We have proved that the new method converges to first-order stationary point with non-asymptotic convergence under some common assumptions. The results of our analysis indicate that our proposed method can achieves a linear speedup in parameter server model.
    \item We have conducted experiments to demonstrate the effectiveness and efficiency of our proposed method on both synthesized and real datasets. The experimental results verify the convergence of the new method and show that our method is able to improve the speedup of distributed training with comparable performance with the state-of-the-art methods. 
\end{itemize}


The rest of the paper is organized as follows. Related works and preliminaries are summarized in Section~\ref{sec:notation_pre}. The detailed {DQGAN} and its convergence rate are described in Section~\ref{sec:method}. The experimental results on both synthetic and real datasets are presented in Section~\ref{sec:experiments}. Conclusions are given in Section~\ref{sec:conclusions}.


\section{Notations and Related Work}
\label{sec:notation_pre}

In this section, we summarize the notations and definitions used
in this paper, and give a brief review of related work.

\subsection{Generative Adversarial Networks}
\label{GANS}

Generative adversarial networks (GANs) consists of two components, one of which is a discriminator ($D$) distinguishing between real data and generated data while the other one is a generator ($G$) generating data to fool the discriminator. We define $p_{r}$ as the real data distribution and $p_{n}$ as the noise probability distribution of the generator $G$. The objective of a GAN is to make the generation data distribution of generator approximates the real data distribution $p_{r}$, which is formulated as a joint loss function for $D$ and $G$~\cite{goodfellow2014generative}
\begin{equation}
    \label{minmax_gan}
    \min_{\theta\in\Theta} \max_{\phi\in\Phi} \mathcal{L} \left( \theta, \phi \right),
\end{equation}
where $\mathcal{L} \left( \theta, \phi \right)$ is defined as follows
\begin{equation}
    \label{loss_gan}
    \begin{aligned} 
        \mathcal{L} \left( \theta, \phi \right) \overset{\text{def}}{=} 
        \mathbb{E}_{\mathbf{x} \sim p_r} \left[\log D_{\phi}\left(\mathbf{x}\right) \right] 
        + \mathbb{E}_{\mathbf{z} \sim p_g} \left[ \log \left( 1-D_{\phi} ( G_{\theta}(\mathbf{z}) ) \right) \right] .
    \end{aligned}
\end{equation}
$D_{\phi}(\mathbf{x})$ indicate that the discriminator which outputs a probability of its input being a real sample. $\theta$ is the parameters of generator $G$ and $\phi$ is the parameters of discriminator $D$. 

However, (\ref{minmax_gan}) may suffer from saturating problem at the early learning stage and leads to vanishing gradients for the generator and inability to train in a stable manner~\cite{goodfellow2014generative,arjovsky2017wasserstein}. Then we used the loss in WGAN~\cite{goodfellow2014generative,arjovsky2017wasserstein}
\begin{equation}
    \label{loss_gan1}
    \begin{aligned} 
        \mathcal{L} \left( \theta, \phi \right) \overset{\text{def}}{=} 
        &  \mathbb{E}_{\mathbf{x} \sim p_r} \left[D_{\phi}\left(\mathbf{x}\right)\right] - \mathbb{E}_{\mathbf{z} \sim p_g} \left[ D_{\phi} ( G_{\theta}(\mathbf{z}) ) \right].
    \end{aligned}
\end{equation}

Then training GAN turns into finding the following Nash equilibrium
\begin{equation}
    \label{G3}
    \theta^{*} \in \mathop{\arg\min}_{\theta\in\Theta} \mathcal{L}_G \left( \theta, \phi^{*} \right),
\end{equation}
\begin{equation}
    \label{D3}
    \phi^{*} \in \mathop{\arg\min}_{\phi\in\Phi} \mathcal{L}_D \left( \theta^{*}, \phi \right),
\end{equation}
where
\begin{equation}
    \label{nonzerosumG}
        \mathcal{L}_G \left( \theta, \phi \right) \overset{\text{def}}{=} 
        -\mathbb{E}_{\mathbf{z} \sim p_{n}} \left[ D_{\phi} (G_{\theta}(\mathbf{z}) )   \right],
\end{equation}
\begin{equation}
    \label{nonzerosumD}
    \begin{aligned} 
        \mathcal{L}_D \left( \theta, \phi \right) \overset{\text{def}}{=} 
         - \mathbb{E}_{\mathbf{x} \sim p_{r}} \left[ D_{\phi}\left(\mathbf{x}\right) \right]  + \mathbb{E}_{\mathbf{z} \sim p_{n}} \left[ D_{\phi} (G_{\theta}(\mathbf{z})   \right].
    \end{aligned}
\end{equation}

Gidel et al.~\cite{gidel2019a} defines a \emph{stationary point} as a couple $(\theta^{*},\phi^{*})$ such that the directional derivatives of both 
$\mathcal{L}_G \left( \theta, \phi^{*} \right)$ and $\mathcal{L}_D \left( \theta^{*}, \phi \right)$ are non-negative, i.e.,
\begin{equation}
    \label{G5}
    \nabla_{\theta} \mathcal{L}_G \left( \theta^{*}, \phi^{*} \right)^{T} \left( \theta-\theta^{*} \right) \geq 0,~\forall(\theta,\phi)\in\Theta\times\Phi;
\end{equation}
\begin{equation}
    \label{D5}
    \nabla_{\phi} \mathcal{L}_D \left( \theta^{*}, \phi^{*} \right)^{T} \left( \phi-\phi^{*} \right) \geq 0,~\forall(\theta,\phi)\in\Theta\times\Phi,
\end{equation}
which can be compactly formulated as
\begin{equation}
    \label{VI}
    \begin{aligned}
        F \left( \mathbf{w}^{*} \right)^{T} & \left( \mathbf{w} - \mathbf{w}^{*} \right) \geq 0, ~ \forall \mathbf{w} \in \Omega,
    \end{aligned}
\end{equation}
where $\mathbf{w} \overset{\text{def}}{=} \left[\theta,~\phi\right]^{T}$, $\mathbf{w}^{*} \overset{\text{def}}{=} \left[\theta^{*},~\phi^{*}\right]^{T}$, $\Omega\overset{\text{def}}{=}\Theta\times\Phi$ and $ F \left( \mathbf{w} \right) \overset{\text{def}}{=}  \left[ \nabla_{\theta} \mathcal{L}_G \left( \theta, \phi \right),~\nabla_{\phi} \mathcal{L}_D \left( \theta, \phi \right) \right]^{T}$.

Many works have been done for GAN. For example, some focus on the loss design, such as WGAN~\cite{arjovsky2017wasserstein}, SN-GAN~\cite{miyato2018spectral}, LS-GAN~\cite{qi2019loss}; while the others focus on the network architecture design, such as CGAN~\cite{mirza2014conditional}, DCGAN~\cite{radford2016unsupervised}, SAGAN~\cite{zhang2018self}. However, only a few works focus on the distributed training of GAN~\cite{liu2019decentralized}, which is our focus.

\subsection{Optimistic Mirror Descent}
\label{OMD}

The update scheme of the basic gradient method is given by 
\begin{equation}
    \label{gradient}
    \mathbf{w}_{t+1} = P_{w} \left[ \mathbf{w}_t - \eta_t F(\mathbf{w}_t) \right],
\end{equation}
where $F(\mathbf{w}_t)$ is the gradient at $\mathbf{w}_t$, $P_{w} \left[\cdot\right]$ is the projection onto the constraint set $w$ (if constraints are present) and $\eta_t$ is the step-size. The gradient descent algorithm is mainly designed for minimization problems and it was proved to be able to converge linearly under the strong monotonicity assumption on the operator $F(\mathbf{w}_t)$~\cite{chen1997Convergence}. However, the basic gradient descent algorithm may produce a sequence that drifts away and cycles without converging when dealing with some special min-max problems~\cite{arjovsky2017wasserstein}, such as bi-linear objective in~\cite{mertikopoulos2019optimistic}. 

To solve the above problem, a practical approach is to compute the average of multiple iterates, which converges with a $O(\frac{1}{\sqrt{t}})$ rate~\cite{nedic2009subgradient}. Recently, the \emph{extragradient} method~\cite{nesterov2007dual} has been used for the min-max problems, due to its superior convergence rate of $O(\frac{1}{t})$. The idea of the \emph{extragradient} can be traced back to Korpelevich~\cite{korpelevich1976extragradient} and Nemirovski~\cite{nemirovski2004prox}. The basic idea of the extragradient is to compute a lookahead gradient to guide the following step. The iterates of the extragradient are given by 

\begin{equation}
    \label{extragradient_11}
    \mathbf{w}_{t+\frac{1}{2}} = P_{w} \left[ \mathbf{w}_t - \eta_t F(\mathbf{w}_t) \right],
\end{equation}
\begin{equation}
    \label{extragradient_12}
    \mathbf{w}_{t+1} = P_{w} \left[ \mathbf{w}_t - \eta_t F(\mathbf{w}_{t+\frac{1}{2}}) \right].
\end{equation}

However, we need to compute the gradient at both $\mathbf{w}_t$ and $\mathbf{w}_{t+\frac{1}{2}}$ in the above iterates. Chiang et al.~\cite{chiang2012online} suggested to use the following iterates

\begin{equation}
    \label{extragradient_21}
    \mathbf{w}_{t+\frac{1}{2}} = P_{w} \left[ \mathbf{w}_t - \eta_t F(\mathbf{w}_{t-\frac{1}{2}}) \right],
\end{equation}
\begin{equation}
    \label{extragradient_22}
    \mathbf{w}_{t+1} = P_{w} \left[ \mathbf{w}_t - \eta_t F(\mathbf{w}_{t+\frac{1}{2}}) \right],
\end{equation}
in which we only need to the compute the gradient at $\mathbf{w}_{t+\frac{1}{2}}$ and reuse the gradient $F(\mathbf{w}_{t-\frac{1}{2}})$ computed in the last iteration.

Considering the unconstrained problem without projection, (\ref{extragradient_21}) and (\ref{extragradient_22}) reduce to 
\begin{equation}
    \label{extragradient_31}
    \mathbf{w}_{t+\frac{1}{2}} = \mathbf{w}_t - \eta_t F(\mathbf{w}_{t-\frac{1}{2}}),
\end{equation}
\begin{equation}
    \label{extragradient_32}
    \mathbf{w}_{t+1} = \mathbf{w}_t - \eta_t F(\mathbf{w}_{t+\frac{1}{2}}),
\end{equation}
and it is easy to see that this update is equivalent to the following one line update as in ~\cite{daskalakis2018training}
\begin{equation}
    \label{omd}
    \mathbf{w}_{t+\frac{1}{2}} = \mathbf{w}_{t-\frac{1}{2}} - 2\eta_t F(\mathbf{w}_{t-\frac{1}{2}})+\eta_t F(\mathbf{w}_{t-\frac{3}{2}}).
\end{equation}

The optimistic mirror descent algorithm is shown Algorithm~\ref{alg:OMD1}. To compute $\mathbf{w}_{t+1}$, it first generates an intermediate state $\mathbf{w}_{t+\frac{1}{2}}$ according to $\mathbf{w}_t$ and the gradient $F(\mathbf{w}_{t-\frac{1}{2}})$ computed in the last iteration, and then computes $\mathbf{w}_{t+1}$ according to both $\mathbf{w}_t$ and the gradient at $\mathbf{w}_{t+\frac{1}{2}}$. In the end of the iteration, $F(\mathbf{w}_{t+\frac{1}{2}})$ is stored for the next iteration. The optimistic mirror descent algorithm was used for online convex optimization~\cite{chiang2012online} and by Rakhlin and general online learning~\cite{rakhlin2013optimization}. Mertikopoulos et al.~\cite{mertikopoulos2019optimistic} used the optimistic mirror descent algorithm for training generative adversarial networks in single machine.

\begin{algorithm}[!h]
    \caption{Optimistic Mirror Descent Algorithm}
    \label{alg:OMD1}
    \begin{algorithmic}[1]
        \REQUIRE {step-size sequence $\eta_t>0$}
        \FOR {$t = 0, 1, \ldots, T-1$}
        \STATE {retrieve $F\left(\mathbf{w}_{t-\frac{1}{2}}\right)$}
        \STATE {set $\mathbf{w}_{t+\frac{1}{2}} = \mathbf{w}_t - \eta_t F(\mathbf{w}_{t-\frac{1}{2}}) $}
        \STATE {compute gradient $F\left(\mathbf{w}_{t+\frac{1}{2}}\right)$ at $\mathbf{w}_{t+\frac{1}{2}}$}
        \STATE {set $\mathbf{w}_{t+1} = P_{w} \left[\mathbf{w}_t - \eta_t F(\mathbf{w}_{t+\frac{1}{2}}) \right]$}
        \STATE {store $F\left(\mathbf{w}_{t+\frac{1}{2}}\right)$}
        \ENDFOR
        \STATE {Return $\mathbf{w}_{T}$}
    \end{algorithmic}
\end{algorithm}

However, all of these works focus on the single machine
setting and we will propose a training algorithm for distributed settings.

\subsection{Distributed Training}

Distributed centralized network and distributed decentralized network are two kinds of topologies used in distributed training.

In distributed centralized network topology, each worker node can obtain information of all other worker nodes. Centralized training systems have different implementations. There exist two common models in the distributed centralized network topology, i.e., the parameter server model~\cite{li2014communication} and the AllReduce model~\cite{rabenseifner2004optimization, wang2019blink}. The difference between the parameter server model and the AllReduce model is that each worker in the AllReduce model directly sends the gradient information to other workers without a server. 


In the distributed decentralized network topology, the structure of computing nodes are usually organized into a graph to process and the network topology does not ensure that each worker node can obtain information of all other worker nodes. All worker nodes can only communicate with their neighbors. In this case, the network topology connection can be fixed or dynamically uncertain. Some parallel algorithms are designed for fixed topology~\cite{jin2016scale, lian2017can, shen2018towards, tang2018d}. On the other hand, the network topology may change when the network accident or power accident causes the connection to be interrupted. Some method have been proposed for this kind of dynamic network topology~\cite{nedic2014distributed, nedic2017achieving}.

Recently, Liu et al.~\cite{liu2019decentralized} proposed to train GAN with the decentralized network topology, but their method suffers from the large amount of parameters exchange problem since they did not compress the transmitted gradients. In this paper, we mainly focus on distributed training of GAN with the distributed centralized network topology.

\begin{figure}[!t]
    \centering
    \includegraphics[width=\columnwidth]{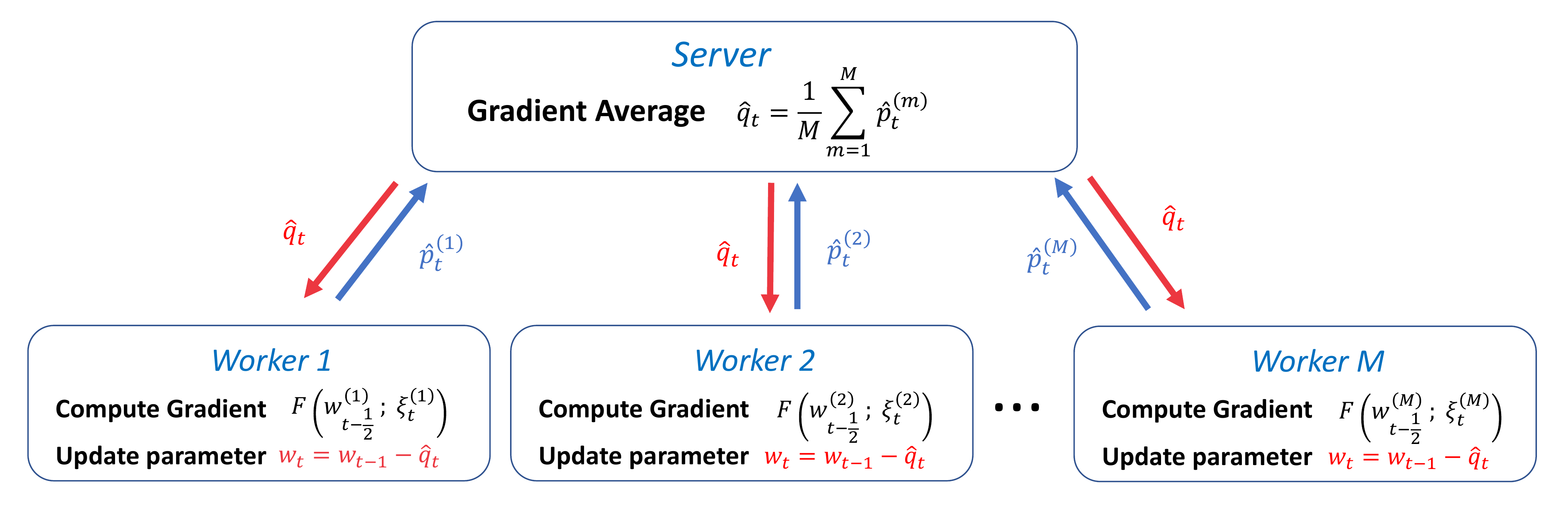}
    \caption{Distributed training of GAN in Algorithm~\ref{alg:DQOSG} in parameter-server model.}
    \label{fig:framework}
\end{figure}

\subsection{Quantization}

Quantization is a promising technique to reduce the neural network model size by reducing the number of bits of the parameters. Early studies on quantization focus on CNNs and RNNs. For example, Courbariaux et al.~\cite{courbariaux2016binarized} proposed to use a single sign function with a scaling factor to binarize the weights and activations in the neural networks, Rastegari et al.~\cite{rastegari2016xnornet} formulated the quantization as an optimization problem in order to quantize CNNs to a binary neural network, Zhou et al.~\cite{zhou2016dorefanet} proposed to quantize the weights, activations and gradients. In the distributed training, the expensive communication cost can be reduced by quantizing the transmitted gradients. 

Denote the quantization function as $Q(\mathbf{v})$ where $\mathbf{v}\in\mathbb{R}^{d}$.
Generally speaking, existing gradient quantization techniques in distributed training can be divided into two categories, i.e., \textbf{biased quantization} ($\mathbb{E}[Q(\mathbf{v})]\neq \mathbf{v}$~for~$\forall\mathbf{v}$) and \textbf{unbiased quantization} ($\mathbb{E}[Q(\mathbf{v})]= \mathbf{v}$ for any $\mathbf{v}$).

\textbf{Biased gradient quantization}: The sign function is a commonly-used biased quantization method~\cite{bernstein2018signsgd,seide20141,strom2015scalable}. Stich et al.~\cite{stich2018sparsified} proposed a top-k quantization method which only retains the top $k$ largest elements of this vector and set the others to zero.

\textbf{Unbiased gradient quantization}: Such methods usually use stochastically quantized gradients~\cite{wen2017terngrad,alistarh2017qsgd,wangni2018gradient}. For example, Alistarh et al.~\cite{alistarh2017qsgd} proposed a compression operator which can be formulated as 
\begin{equation}
\label{qsgd}
    Q(v_i) = sign(v_i) \cdot \|\mathbf{v}\|_{2} \cdot \xi_i(\mathbf{v},s),
\end{equation}
where $\xi_i(\mathbf{v},s)$ is defined as follows
\begin{equation}
    \xi_i(\mathbf{v},s)=
    \left\{  
        \begin{aligned}  
            &\frac{l}{s} &\text{with prob.} ~ 1-\left(\frac{|v_i|}{\|\mathbf{v}\|_2}\cdot s - l\right); \\  
            &\frac{l+1}{s} &\text{otherwise}.
        \end{aligned},
    \right.
\end{equation}
where $s$ is the number of quantization levels, $0 \leq l < s$ is a integer such that $|v_i|/\|\mathbf{v}\|_{2}\in[l/s,(l+1)/s]$. Hou et al.~\cite{hou2018analysis} replaced the $\|\mathbf{v}\|_{2}$ in the above method with $\|\mathbf{v}\|_{\infty}$.



\section{The Proposed Method}
\label{sec:method}

\subsection{{DQGAN} Algorithm}

Actually, we consider extensions of the algorithm to the context of a \emph{stochastic} operator, i.e., we no longer have access to the exact gradient but to an unbiased stochastic estimate of it.
Suppose we have $M$ machines.
When in machine $m$ at $t$-th iteration, we sample a mini-batch according to $\xi_{t}^{(m)} = \left( \xi_{t,1}^{(m)}, \xi_{t,2}^{(m)}, \cdots, \xi_{t,B}^{(m)} \right)$, where $B$ is the minibatch size. 
In all $M$ machines, we use the same mini-batch size $B$.
We use the term $F(\mathbf{w}_{t}^{(m)};\xi_{t,b}^{(m)})$ ~ ($1\leq b \leq B$) and $F(\mathbf{w}_{t}^{(m)})$ to stand for \emph{stochastic gradient} and \emph{gradient} respectively.
When in machine $m$ at $t$-th iteration, we define \emph{mini-batch gradient} as $F(\mathbf{w}_{t}^{(m)};\xi_{t}^{(m)}) = \frac{1}{B} \sum_{b=1}^{B} F(\mathbf{w}_{t}^{(m)};\xi_{t,b}^{(m)})$. 

To reduce the size of the transmitted gradients in distributed training, we introduce a quantization function $Q(\cdot)$ to compress the gradients. In this paper, we consider a general $\delta$-approximate compressor for our method
\begin{defn}
    An quantization operator $Q$ is said to be $\delta$-approximate compressor for $\delta \in (0,1])$ if 
    \begin{equation}
        \| Q\left(F(\mathbf{w})\right)-F(\mathbf{w}) \|^{2} \leq (1-\delta) \|F(\mathbf{w})\|^{2} \quad \text{for all} ~ \mathbf{w} \in \Omega.
    \end{equation}
    \label{definQ}
\end{defn}
\vskip -0.2in
Before sending gradient data to the central server, each worker needs quantify the gradient data, which makes our algorithm have a small communication overhead. 
In addition, our algorithm is applicable to any gradient compression method that satisfies a general $\delta$-approximate compressor.

In general, the quantized gradient will lose some accuracy and cause the algorithm to fail to converge. In this paper, we have designed an error compensation operation to solve this problem, which is to incorporate the error made by the compression operator into the next step to compensate for gradients. Recently, Stich et al.~\cite{stich2018sparsified} conducted the theoretical analysis of error-feedback in the strongly convex case and Karimireddy et al.~\cite{karimireddy2019error} further extended the convergence results to the non-convex and weakly convex cases. 
These error-feedback operations are designed to solve the minimization problem. Here, in order to solve the min-max problem, we need to design the error-feedback operation more cleverly. In each iteration, we compensate the error caused by gradient quantization twice.

The proposed method, named as {Distributed Quantized Generative Adversarial Networks (DQGAN)}, is summarized in Algorithm~\ref{alg:DQOSG}. The procedure of the new method is illustrated with the the parameter server model~\cite{li2014communication} in Figure~\ref{fig:framework}. 
There are $M$ works participating in the model training. 
In this method, $\mathbf{w}_{0}$ is first initialized and pushed to all workers, and the local variable $\mathbf{e}^{m}_{0}$ is set to $0$ for $m\in[1,M]$. 

In each iteration, the state $\mathbf{w}_{t-1}^{(m)}$ is updated to obtain the intermediate state $\mathbf{w}_{t-\frac{1}{2}}^{(m)}$ on $m$-th worker. Then each worker computes the gradients and adds the error compensation. The $m$-th worker computes the gradients $F(\mathbf{w}_{t-\frac{1}{2}}^{(m)};\xi_{t}^{(m)})$ and adds the error compensation $\mathbf{e}^{(m)}_{t-1}$ to obtain $\mathbf{p}_{t}^{(m)}$, and then quantizes it to $\hat{\mathbf{p}}_{t}^{(m)}$ and transmit the quantized one to the server. The error $\mathbf{e}_{t}^{(m)}$ is computed as the difference between $\mathbf{p}_{t}^{(m)}$ and $\hat{\mathbf{p}}_{t}^{(m)}$. The server will average all quantized gradients and push it to all workers. Then the new parameter $\mathbf{w}_{t}$ will be updated.
Then we use it to update $\mathbf{w}_{t-1}$ to get the new parameter $\mathbf{w}_{t}$.

\begin{algorithm}[!h]
    \caption{The Algorithm of {DQGAN}}
    \begin{algorithmic}[1]
    \REQUIRE {step-size $\eta>0$, Quantization function $Q\left(\cdot\right)$} is $\delta$-approximate compressor. 
    \STATE Initialize $\mathbf{w}_0$ and push it to all workers, set $\mathbf{w}^{(m)}_{-\frac{1}{2}} = \mathbf{w}_0$ and $\mathbf{e}^{(m)}_0=0$ for $1 \leq m \leq M$.
    \FOR {$t = 1, 2, \ldots, T$}
    \STATE {\textbf{on} worker m : $( m \in \{1, 2, \cdots, M\} )$}
    \STATE {\quad set $\mathbf{w}_{t-\frac{1}{2}}^{(m)} = \mathbf{w}_{t-1} - \left[ \eta F(\mathbf{w}_{t-\frac{3}{2}}^{(m)};\xi_{t-1}^{(m)}) + \mathbf{e}^{(m)}_{t-1} \right]$}
    \STATE {\quad compute gradient $F(\mathbf{w}_{t-\frac{1}{2}}^{(m)};\xi_{t}^{(m)})$}
    \STATE {\quad set $\mathbf{p}_{t}^{(m)} = \eta F(\mathbf{w}_{t-\frac{1}{2}}^{(m)};\xi_{t}^{(m)}) + \mathbf{e}^{(m)}_{t-1} $}
    \STATE {\quad set $\hat{\mathbf{p}}_{t}^{(m)} = Q\left( \mathbf{p}_{t}^{(m)} \right)$} and push it to the server
    \STATE {\quad set $\mathbf{e}_{t}^{(m)} = \mathbf{p}_{t}^{(m)} - \hat{\mathbf{p}}_{t}^{(m)}$}
    \STATE {\textbf{on} server:}
    \STATE {\quad \textbf{pull} $\hat{\mathbf{p}}_{t}^{(m)}$ \textbf{from} each worker}
    \STATE {\quad set $\hat{\mathbf{q}}_{t} = \frac{1}{M} \left[ \sum_{m=1}^{M} \hat{\mathbf{p}}_{t}^{(m)} \right]$}
    \STATE {\quad \textbf{push} $\hat{\mathbf{q}}_{t}$ \textbf{to} each worker}
    \STATE {\textbf{on} worker m : $( m \in \{1, 2, \cdots, M\} )$}
    \STATE {\quad set $\mathbf{w}_{t} = \mathbf{w}_{t-1} - \hat{\mathbf{q}}_{t}$}
    \ENDFOR
    \STATE {Return $\mathbf{w}_{T}$}
    \end{algorithmic}
    \label{alg:DQOSG}
\end{algorithm}

\subsection{Coding Strategy}

The quantization plays an important role in our method. In this paper, we proposed a general $\delta$-approximate compressor in order to include a variety of quantization methods for our method. In this subsection, we will prove that some commonly-used quantization methods are $\delta$-approximate compressors.

According to the definition of the $k$-contraction operator~\cite{stich2018sparsified}, we can verify the following theorem
\begin{thm}
    The $k$-contraction operator~\cite{stich2018sparsified} is a $\delta$-approximate compressor with $\delta=\frac{k}{d}$ where $d$ is the dimensions of input and $k\in(0,d]$ is a parameter.
    \label{deltaOne}
\end{thm}

Moreover, we can prove the following theorem

\begin{thm}
    The quantization methods in~\cite{alistarh2017qsgd} and~\cite{hou2018analysis} are $\delta$-approximate compressors.
    \label{deltaTwo}
\end{thm}

Therefore, a variety of quantization methods can be used for our method.

\subsection{Convergence Analysis}
Throughout the paper, we make the following assumption

\begin{ass}[Lipschitz Continuous]
    \begin{enumerate}
        \item $F$ is $L$-Lipschitz continuous, i.e. 
        $\| F(\mathbf{w}_{1}) - F(\mathbf{w}_{2}) \| \leq L \| \mathbf{w}_{1} - \mathbf{w}_{2} \|$ for $\forall \mathbf{w}_{1}, \mathbf{w}_{2}$ .
        \item $\| F(\mathbf{w}) \| \leq G$ for $\forall \mathbf{w}$ .
    \end{enumerate}
    \label{assum1}
\end{ass}

\begin{ass}[Unbiased and Bounded Variance]
    For $\forall \mathbf{w}_{t-\frac{1}{2}}^{(m)}$, we have $\mathbb{E}\left[F(\mathbf{w}_{t-\frac{1}{2}}^{(m)};\xi_{t,b}^{(m)})\right] = F(\mathbf{w}_{t-\frac{1}{2}}^{(m)})$ and $\mathbb{E}\|F(\mathbf{w}_{t-\frac{1}{2}}^{(m)};\xi_{t,b}^{(m)}) - F(\mathbf{w}_{t-\frac{1}{2}}^{(m)})\|^2 \leq \sigma^2$ , where $1 \leq b \leq B$.
    \label{assum2}
\end{ass}

\begin{ass}[Pseudomonotonicity] 
    The operator $F$ is pseudomonotone, i.e., 
    $$\left\langle F(\mathbf{w}_{2}), \mathbf{w}_{1} - \mathbf{w}_{2} \right\rangle \geq 0 \Rightarrow \left\langle F(\mathbf{w}_{1}) , \mathbf{w}_{1} - \mathbf{w}_{2} \right\rangle \geq 0 \quad \text{for} \quad \forall \mathbf{w}_{1},\mathbf{w}_{2}$$
    \label{assum3}
\end{ass}

Now, we give a key Lemma, which shows that the residual errors maintained in Algorithm~\ref{alg:DQOSG} do not accumulate too much. 

\begin{lem}
    At any iteration $t$ of Algorithm~\ref{alg:DQOSG}, the norm of the error $\frac{1}{M}\sum_{m=1}^{M}\| \mathbf{e}_{t}^{(m)} \|^{2}$ is bounded: 
    \begin{equation}
        \mathbb{E} \left[ \| \frac{1}{M} \sum_{m=1}^{M} \mathbf{e}^{(m)}_{t} \|^2 \right] 
        \leq \frac{8\eta^2(1-\delta)(G^{2}+\frac{\sigma^2}{B})}{\delta^2} 
    \end{equation}
    If $\delta = 1$, then $\|\mathbf{e}_{t}^{(m)}\|=0$ for $1 \leq m \leq M$ and the error is zero as expected.
    \label{errorBounded}
\end{lem}

Finally, based on above assumptions and lemma, we are on the position to describe the convergence rate of Algorithm~\ref{alg:DQOSG}.

\begin{thm}
    By picking the $\eta \leq \min \{ \frac{1}{\sqrt{BM}}, \frac{1}{6\sqrt{2}L} \}$ in Algorithm~\ref{alg:DQOSG}, we have 
    \begin{equation}
        \begin{aligned}
            \frac{1}{T} \sum_{t=1}^{T} \mathbb{E} & \left[ \| \frac{1}{M}\sum_{m=1}^{M} F\left(\mathbf{w}_{t-\frac{1}{2}}^{(m)};\xi_{t}^{(m)}\right) \|^2 \right] 
            \leq \frac{4 \| \tilde{\mathbf{w}}_{0} - \mathbf{w}^* \|^2}{\eta^{2} T} + \frac{1728~ L^2 \sigma^2}{B^2M^2} \\
            & \quad + \frac{3456~ L^2 G^2 (M-1)}{BM^2} + \frac{9216~ L^2 (1-\delta)(G^{2}+\frac{\sigma^2}{B})(M-1)}{\delta^2 BM^2} + \frac{48~ \sigma^2}{BM}
        \end{aligned}
        \label{eq:converg}
    \end{equation}
    \label{converge}
\end{thm}

Theorem~\ref{converge} gives such a non-asymptotic convergence and linear speedup in theory. 
We need to find the $\epsilon$-first-order stationary point, i.e., 
\begin{equation}
    \mathbb{E} \left[ \| \frac{1}{M}\sum_{m=1}^{M} F\left(\mathbf{w}_{t-\frac{1}{2}}^{(m)};\xi_{t}^{(m)}\right) \|^2 \right] \leq \epsilon^{2}
\end{equation}
By taking $M=\mathcal{O}(\epsilon^{-2})$ and $T=\mathcal{O}(\epsilon^{-8})$, 
we can guarantee that the algorithm~\ref{alg:DQOSG} can reach an $\epsilon$-first-order stationary point. 



\section{Experiments}
\label{sec:experiments}

\begin{figure}[t]
    \centerline{\includegraphics[width=\columnwidth]{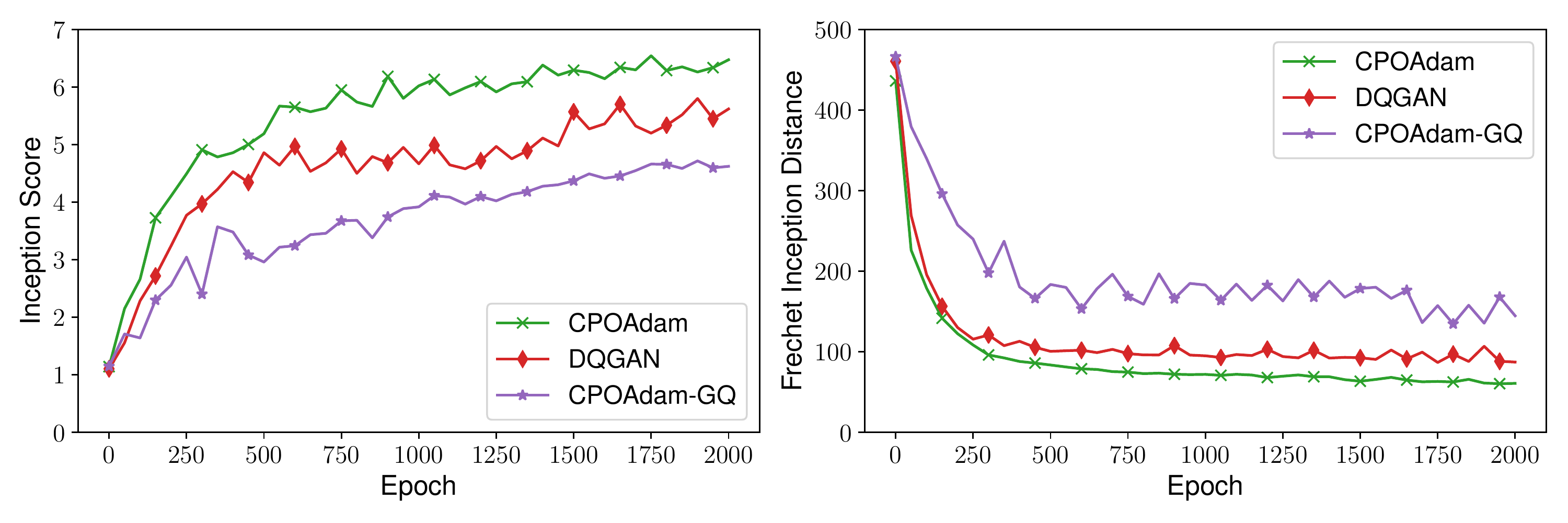}}
    \caption{The Inception Score and Fr{\'e}chet Inception Distance values of three methods on the CIFAR10 dataset.}
    \label{fig:results_cifa10}
    \vspace{-0.4cm}
\end{figure}

\begin{figure}[t]
   \centerline{\includegraphics[width=\columnwidth]{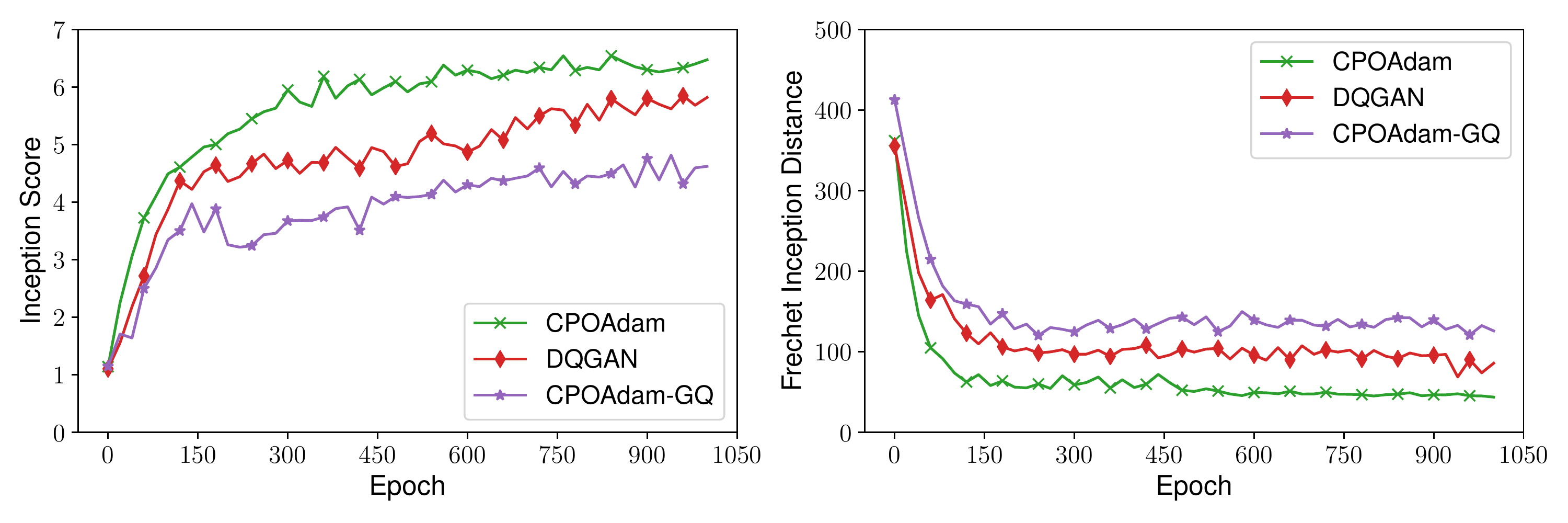}}
    \caption{The Inception Score and Fr{\'e}chet Inception Distance values of three methods on the CelebA dataset.}
    \label{fig:results_celeba}
    \vspace{-0.4cm}
\end{figure}

\begin{figure}[t]
    \centerline{\includegraphics[width=0.8\columnwidth]{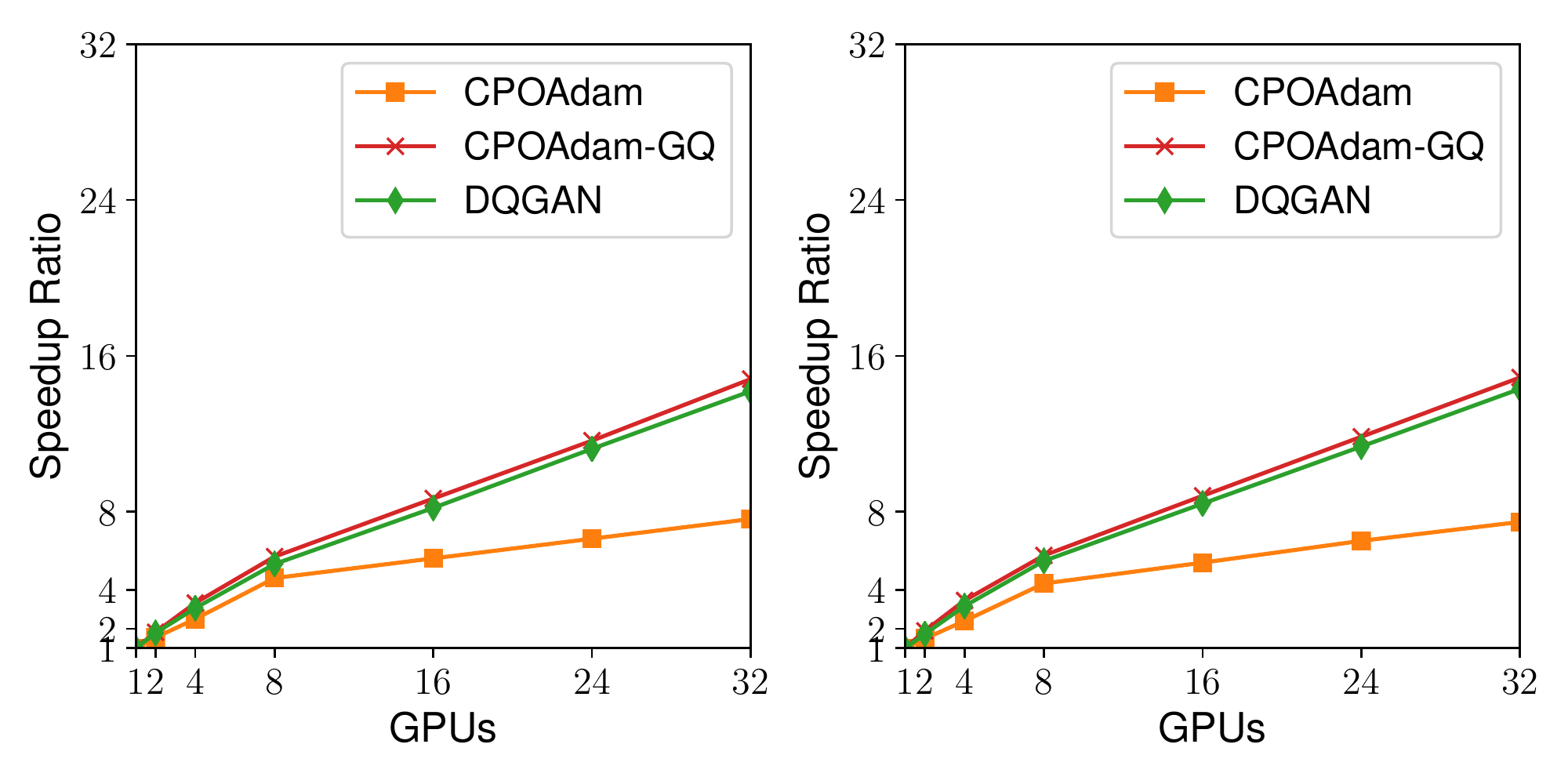}}
    \caption{The speedup of our method on the CIFAR10 and CelebA datasets (left and right respectively).}
    \label{fig:speedup}
    \vspace{-0.4cm}
\end{figure}

In this section, we present the experimental results on 
real-life datasets. We used PyTorch~\cite{paszke2019pytorch} as the underlying deep learning framework 
and Nvidia NCCL~\cite{nccl} as the communication mechanism.
We used the following two real-life benchmark datasets for the remaining experiments. 
The \textbf{CIFAR-10} dataset~\cite{krizhevsky2009learning} contains 60000 32x32 images in 10 classes: airplane, automobile, bird, cat, deer, dog, frog, horse, ship and truck. 
The \textbf{CelebA} dataset~\cite{liu2015faceattributes} is a large-scale face attributes dataset with more than 200K celebrity images, each with 40 attribute annotations. 

We compared our method with the 
{Centralized Parallel Optimistic Adam (CPOAdam)} which is our method without quantization and error-feedback, 
and the Centralized Parallel Optimistic Adam with Gradients Quantization (CPOAdam-GQ) 
for training the generative adversarial networks with the loss in (\ref{loss_gan1}) and 
the deep convolutional generative adversarial network (DCGAN) architecture~\cite{radford2016unsupervised}. 
We implemented full-precision (float32) baseline models and set the number of bits for our method to $8$. 
We used the compressor in~\cite{hou2018analysis} for our method. 
Learning rates and hyperparameters in these methods were chosen by an inspection of grid search results 
so as to enable a fair comparison of these methods. 
We use the Inception Score~\cite{salimans2016improved} and fr{\'e}chet inception distance~\cite{dowson1982frechet} 
to evaluate all methods, where higher inception score and lower Fr{\'e}chet Inception Distance indicate better results.

The Inception Score and Fréchet Inception Distance values of three methods on the CIFAR10 and CelebA datasets 
are shown in Figures~\ref{fig:results_cifa10} and \ref{fig:results_celeba}. 
The results on both datasets show that the CPOAdam achieves high inception scores and 
lower fr{\'e}chet inception distances with all epochs of training.
We also notice that our method with 1/4 full precision gradients is able to produce comparable result 
with CPOAdam with full precision gradients, 
finally with at most $0.6$ decrease in terms of Inception Score and 
at most $30$ increase in terms of Fr{\'e}chet Inception Distance on the CIFAR10 dataset, 
with at most $0.5$ decrease in terms of Inception Score and 
at most $40$ increase in terms of Fr{\'e}chet Inception Distance on the CelebA dataset.


Finally, we show the speedup results of our method on the CIFAR10 and CelebA datasets in Figure~\ref{fig:speedup}, 
which shows that our method is able to improve the speedup of training GANs and 
the improvement becomes more significant with the increase of data size. 
With 32 workers, our method with $8$ bits of gradients 
achieves a significant improvement
on the CIFAR10 and the CelebA dataset, compared to the CPOAdam.
This is the same as our expectations, after all, 8bit transmits less data than 32bit.


\section{Conclusion}
\label{sec:conclusions}

In this paper, we have proposed a distributed optimization algorithm for training GANs. 
The new method reduces the communication cost via gradient compression, 
and the error-feedback operations we designed is used to compensate for the bias caused 
by the compression operation and ensure the non-asymptotic convergence. 
We have theoretically proved the non-asymptotic convergence of the new method, 
with the introduction of a general $\delta$-approximate compressor. 
Moreover, we have proved that the new method has linear speedup in theory. 
The experimental results show that our method is able to produce comparable results as the distributed OMD without quantization, 
only with slight performance degradation.

Although our new method has linear speedup in theory, 
the cost of gradients synchronization affects its performance in practice. 
Introducing gradient asynchronous communication technology will break the synchronization barrier and 
improve the efficiency of our method in real applications. We will leave this for the future work.

\bibliography{arxiv}
\bibliographystyle{plain}

%
%

\appendix
\newpage

\section{The Proof of Theorem~\ref{deltaOne} and Theorem~\ref{deltaTwo}}

According to the definition of the $k$-contraction operator
, we can verify the Theorem~\ref{deltaOne}. 

\begin{proof}
    The $k$-contraction operator is a operator \emph{comp}: $\mathbb{R}^{d} \rightarrow \mathbb{R}^{d}$ that satisfies the contraction property
    \begin{equation}
        \mathbb{E} \| \mathbf{v} - \text{comp}(\mathbf{v}) \|^2 \leq \left( 1 - \frac{k}{d} \right) \| \mathbf{v} \|^2 
        ~ \forall \mathbf{v} \in \mathbb{R}^{d},
    \end{equation}
    where $k$ is a parameter that satisfies $0 < k \leq d$. 
    
    We set $\delta = \frac{k}{d}$. Then we have $\delta \in (0, 1]$. Obviously, the $k$-contraction operator is a $\delta$-approximate compressor with $\delta=\frac{k}{d}$. 
\end{proof}

For example, we can directly take the largest $k~(0 < k \leq d)$ elements of $d$ elements in $\mathbf{v}$ in practice.

Moreover, we can prove the Theorem~\ref{deltaTwo}. 

\begin{proof}
    The $m$-bit stochastic quantification method 
    is defined as 
    \begin{equation}
        Q(v_i) = s \cdot \text{sign} (v_i) \cdot q(v_i, s) ,
    \end{equation}
    where $v_i$ is the $i$-th element of vector $\mathbf{v}$, 
    $s$ is a scalar and equals $\|\mathbf{v}\|_{2}$ 
    or $\|\mathbf{v}\|_{\infty}$ 
    , 
    $q(v_i, s) \in \{ -B_k, \cdots, -B_1, B_0, B_1, \cdots, B_k \}$ and $0 = B_0 < B_1 < \cdots < B_k$,
    $k = 2^{m-1} - 1$ denotes the number of quantized values that $m$-bit can represent. 
    $q(v_i, s)$ is defined as 
    \begin{equation}
        q(v_i, s) = 
        \left\{  
            \begin{aligned}  
                &B_{r+1} &\text{with prob.} ~ \frac{\frac{|v_i|}{s} - B_{r}}{B_{r+1} - B_{r}}; \\  
                &B_{r} &\text{otherwise}.
            \end{aligned},
        \right.
    \end{equation}
    where $r \in \{0, 1, \cdots, k-1\}$ satisfies $B_{r} \leq \frac{|v_i|}{s} < B_{r+1}$. 

    So we have 
    \begin{equation}
        \begin{aligned}
            \mathbb{E} \left[ Q(v_i) \right] 
            = s \cdot \text{sign} (v_i) \cdot 
            \left( B_{r+1} \cdot \frac{\frac{|v_i|}{s} - B_{r}}{B_{r+1} - B_{r}} + 
            B_{r} \cdot \frac{B_{r+1} - \frac{|v_i|}{s}}{B_{r+1} - B_{r}} \right) 
            = v_i
        \end{aligned}
    \end{equation}
    Note that $Q(\mathbf{v})$ is an unbiased estimator of $\mathbf{v}$.

    To prove that the quantization methods 
    are $\delta$-approximate compressors, 
    i.e.,  
    \begin{equation}
        \mathbb{E} \left[ \| Q(\mathbf{v}) - \mathbf{v} \|^2 \right] \leq (1 - \delta) \| \mathbf{v} \|^2 ,
    \end{equation}
    we just need to prove that for all element in $\mathbf{v}$, $Q(\cdot)$ satisfies 
    \begin{equation}
        \mathbb{E} \left[ | Q(v_i) - v_i |^2 \right] \leq (1 - \delta) | v_i |^2 ,
        \label{eq:vigoal}
    \end{equation}
    where $v_i$ is the $i$-th element of vector $\mathbf{v}$. 
    When $v_i = 0$, $Q(v_i) = 0$ and formula~\eqref{eq:vigoal} holds. So our analysis below only considers the case of $v_i \neq 0$.

    Since 
    \begin{equation}
        \begin{aligned}
            \mathbb{E} \left[ \| Q(v_i) \|^2 \right] 
            &= s^2 \cdot \left( B_{r+1}^2 \cdot \frac{\frac{|v_i|}{s} - B_{r}}{B_{r+1} - B_{r}} + 
            B_{r}^2 \cdot \frac{B_{r+1} - \frac{|v_i|}{s}}{B_{r+1} - B_{r}} \right) \\ 
            &= s^2 \cdot \frac{\frac{|v_i|}{s} \left( B_{r+1}^2 - B_{r}^2 \right) 
            - B_{r}B_{r+1} \left( B_{r+1} - B_{r} \right) }{B_{r+1} - B_{r}} \\ 
            &= |v_i| \cdot s \cdot \left( B_{r+1} + B_{r} \right) - B_{r}B_{r+1} \cdot s^2 \\ 
        \end{aligned}
    \end{equation}
    and $\mathbb{E} \left[ | Q(v_i) - v_i |^2 \right] = \mathbb{E} \left[ | Q(v_i) |^2 \right] - | v_i |^2$ , 
    the goal we need to prove is changed to 
    \begin{equation}
        |v_i| \cdot s \cdot \left( B_{r+1} + B_{r} \right) - B_{r}B_{r+1} \cdot s^2 \leq (2 - \delta) | v_i |^2 .
    \end{equation}

    We set that $C_{r} = s \cdot B_{r}$, $C_{r+1} = s \cdot B_{r+1}$ and $a = |v_i|$ satisfies $C_{r} \leq a < C_{r+1}$. 
    
    Now, our goal is that we can find $\delta \in (0, 1]$ satisfying 
    \begin{equation}
        (2 - \delta) \cdot a^2 - \left( C_{r} + C_{r+1} \right) \cdot a + C_{r} C_{r+1} \geq 0 \quad \forall ~ C_{r} \leq a < C_{r+1}.
        \label{eq:newgoal}
    \end{equation}

    This is easy to prove and we provide a proof method below.

    After rearranging~\eqref{eq:newgoal}, there is
    \begin{equation}
        \delta \leq 2 - \frac{\left( C_{r} + C_{r+1} \right) \cdot a - C_{r} C_{r+1}}{a^2} \quad \forall ~ C_{r} \leq a < C_{r+1}.
    \end{equation}
    So, we just need to prove that 
    \begin{equation}
        0 < 2 - \frac{\left( C_{r} + C_{r+1} \right) \cdot a - C_{r} C_{r+1}}{a^2} \leq 1 \quad \forall ~ C_{r} \leq a < C_{r+1}.
        \label{eq:newgoal2}
    \end{equation}
    As long as~\eqref{eq:newgoal2} holds, $\delta$ that satisfies $\delta \in (0, 1]$ exists.
    
    First, we prove that 
    \begin{equation}
        2 - \frac{\left( C_{r} + C_{r+1} \right) \cdot a - C_{r} C_{r+1}}{a^2} > 0 \quad \forall ~ C_{r} \leq a < C_{r+1}.
    \end{equation}
    That is 
    \begin{equation}
        2 a^2 - \left( C_{r} + C_{r+1} \right) \cdot a + C_{r} C_{r+1} > 0 \quad \forall ~ C_{r} \leq a < C_{r+1}.
    \end{equation}
    For this one-variable quadratic inequality, we just prove 
    \begin{equation}
        \frac{C_{r} + C_{r+1} + \sqrt{\left( C_{r} + C_{r+1} \right)^2 - 8 C_{r} C_{r+1}}}{4} < C_{r} .
        \label{eq:onegoal}
    \end{equation}
    \eqref{eq:onegoal} obviously holds after rearranging as follows. 
    \begin{equation}
        \sqrt{\left( C_{r} + C_{r+1} \right)^2 - 8 C_{r} C_{r+1}} < 3 C_{r} - C_{r+1} .
    \end{equation}

    Second, we prove that 
    \begin{equation}
        2 - \frac{\left( C_{r} + C_{r+1} \right) \cdot a - C_{r} C_{r+1}}{a^2} \leq 1 \quad \forall ~ C_{r} \leq a < C_{r+1}.
    \end{equation}
    That is 
    \begin{equation}
        a^2 - \left( C_{r} + C_{r+1} \right) \cdot a + C_{r} C_{r+1} \leq 0 \quad \forall ~ C_{r} \leq a < C_{r+1}.
    \end{equation}
    After rearranging, 
    \begin{equation}
        \left( a - C_{r} \right) \left( a - C_{r+1} \right) \leq 0 \quad \forall ~ C_{r} \leq a < C_{r+1}. 
        \label{eq:twogoal}
    \end{equation}
    \eqref{eq:twogoal} obviously holds. 
    So far, theorem~\ref{deltaTwo} has been proved.
\end{proof}

\section{Proof of Lemma and Theorem}


\subsection{Proof of Lemma~\ref{errorBounded}}
\begin{proof}
    By definition of the error sequence, 
    \begin{equation}
        \begin{aligned}
            & \| \mathbf{e}^{(m)}_{t} \|^2 = \| \mathbf{p}^{(m)}_{t} - Q(\mathbf{p}^{(m)}_{t}) \|^2 
                \leq (1-\delta) \| \mathbf{p}^{(m)}_{t} \|^2 \\
                &= (1-\delta) \| \eta F(\mathbf{w}_{t-\frac{1}{2}}^{(m)};\xi_{t}^{(m)}) + \mathbf{e}^{(m)}_{t-1} \|^2 \\ 
                &= (1-\delta)\|\mathbf{e}^{(m)}_{t-1}\|^2 + (1-\delta)\eta^{2}\| F(\mathbf{w}_{t-\frac{1}{2}}^{(m)};\xi_{t}^{(m)}) \|^2 
                + (1-\delta) \left\langle \eta F(\mathbf{w}_{t-\frac{1}{2}}^{(m)};\xi_{t}^{(m)}), \mathbf{e}^{(m)}_{t-1} \right\rangle \\ 
                &\leq (1-\delta)\left(1+K_1\right)\|\mathbf{e}^{(m)}_{t-1}\|^2 
                + (1-\delta)\left(1+\frac{1}{K_1}\right)\eta^{2}\|F(\mathbf{w}_{t-\frac{1}{2}}^{(m)};\xi_{t}^{(m)})\|^2
        \end{aligned}
    \end{equation}
    In the last inequality, we use the Young's inequality and $K_1 > 0$ is constant. 
    So, we give the simple algebraic computations to solve the recurrence relation above:
    \begin{equation}
        \begin{aligned}
            &\mathbb{E} \left[ \| \mathbf{e}^{(m)}_{t} \|^2 \right] \\
            &\leq (1-\delta)\left(1+K_1\right)\mathbb{E}\left[\|\mathbf{e}^{(m)}_{t-1}\|^2\right] 
            + (1-\delta)\left(1+\frac{1}{K_1}\right) 
             \cdot \mathbb{E}\left[ \eta^{2}\|F(\mathbf{w}_{t-\frac{1}{2}}^{(m)};\xi_{t}^{(m)})\|^2 \right] \\ 
            &\leq (1-\delta)\left(1+K_1\right)\mathbb{E}\left[\|\mathbf{e}^{(m)}_{t-1}\|^2\right] \\
            & \quad + (1-\delta)\left(1+\frac{1}{K_1}\right) \eta^{2} 
             \cdot \mathbb{E}\left[ 2 \|F(\mathbf{w}_{t-\frac{1}{2}}^{(m)})\|^2 + 2 \|F(\mathbf{w}_{t-\frac{1}{2}}^{(m)}) - F(\mathbf{w}_{t-\frac{1}{2}}^{(m)};\xi_{t}^{(m)})\|^2 \right] \\ 
            &\leq \sum^{t}_{i=0} \left[ (1-\delta)\left(1+K_1\right) \right]^{t-i} (1-\delta)\left(1+\frac{1}{K_1}\right) 
            2 \eta^2 (G^{2} + \frac{\sigma^2}{B}) \\ 
            &\leq \sum^{\infty}_{i=0} \left[(1-\delta)\left(1+K_1\right)\right]^{i} (1-\delta)\left(1+\frac{1}{K_1}\right) 
            2 \eta^2 (G^{2} + \frac{\sigma^2}{B}) \\ 
            &= \frac{(1-\delta)\left(1+\frac{1}{K_1}\right)}{1-\left[(1-\delta)\left(1+K_1\right)\right]} 2 \eta^2 (G^{2} + \frac{\sigma^2}{B}) \\ 
            &= \frac{(1-\delta)\left(1+\frac{1}{K_1}\right)}{\delta-K_1\left(1-\delta\right)} 2 \eta^2 (G^{2} + \frac{\sigma^2}{B})
        \end{aligned}
    \end{equation}
    We take $K_1 = \frac{\delta}{2(1-\delta)}$ such that $1 + \frac{1}{K_1} = \frac{2-\delta}{\delta} \leq \frac{2}{\delta}$.
    Plugging this in the above gives
    \begin{equation}
        \begin{aligned}
            \mathbb{E} \left[ \| \mathbf{e}^{(m)}_{t} \|^2 \right] 
            &\leq \frac{(1-\delta)\left(1+\frac{1}{K_1}\right)}{\delta-K_1\left(1-\delta\right)} \cdot 2 \eta^2 (G^{2} + \frac{\sigma^2}{B}) \\
            &= \frac{2(1-\delta)\left(1+\frac{1}{K_1}\right)}{\delta} \cdot 2 \eta^2 (G^{2} + \frac{\sigma^2}{B}) \\
            &\leq \frac{8\eta^2(1-\delta)(G^{2}+\frac{\sigma^2}{B})}{\delta^2} 
        \end{aligned}
    \end{equation}
    Then we have
    \begin{equation}
        \begin{aligned}
            \mathbb{E} \left[ \| \frac{1}{M} \sum_{m=1}^{M} \mathbf{e}^{(m)}_{t} \|^2 \right]
                &= \mathbb{E} \left[ \frac{1}{M^2} \| \sum_{m=1}^{M} \mathbf{e}^{(m)}_{t} \|^2 \right] \\ 
                &\leq \mathbb{E} \left[ \frac{1}{M^2} \cdot M \sum_{m=1}^{M} \| \mathbf{e}^{(m)}_{t} \|^2 \right] \\ 
                &= \mathbb{E} \left[ \frac{1}{M} \cdot \sum_{m=1}^{M} \| \mathbf{e}^{(m)}_{t} \|^2 \right] \\ 
                &\leq \frac{8\eta^2(1-\delta)(G^{2}+\frac{\sigma^2}{B})}{\delta^2}.
        \end{aligned}
    \end{equation} 
    In the first inequality, we used the Cauchy-Schwarz inequality. 
\end{proof}

\vspace*{0.1in}
\subsection{Main Proof of Theorem of~\ref{converge}}
\begin{proof}
    
    Define $\bar{\mathbf{w}}_{t-\frac{1}{2}} = \frac{1}{M}\sum_{m=1}^{M} \mathbf{w}_{t-\frac{1}{2}}^{(m)}$, $\mathbf{\epsilon}_{t}^{(m)} = F\left(\mathbf{w}_{t-\frac{1}{2}}^{(m)};\xi_{t}^{(m)}\right) - F\left(\mathbf{w}_{t-\frac{1}{2}}^{(m)}\right)$ and $\bar{\mathbf{\epsilon}}_{t} = \frac{1}{M} \sum_{m=1}^{M} \mathbf{\epsilon}_{t}^{(m)}$ .
    
    We consider the error-corrected sequence $\tilde{\mathbf{w}}_t$ which represents $\mathbf{w}_t$ 
    with the \emph{delayed} information added: 
    \begin{equation}
        \tilde{\mathbf{w}}_t = \mathbf{w}_{t} - \frac{1}{M}\sum_{m=1}^{M}\mathbf{e}_{t}^{(m)} . 
    \end{equation}
    It satisfies the recurrence
    \begin{equation}
        \label{eq:tilde}
        \begin{aligned}
            \tilde{\mathbf{w}}_{t} 
            &= \mathbf{w}_{t} - \frac{1}{M}\sum_{m=1}^{M}\mathbf{e}_{t}^{(m)} \\
            &= \mathbf{w}_{t-1} - \frac{1}{M}\sum_{m=1}^{M} Q\left( \mathbf{p}_{t}^{(m)} \right) - \frac{1}{M}\sum_{m=1}^{M}\mathbf{e}_{t}^{(m)} \\ 
            &= \mathbf{w}_{t-1} - \frac{1}{M}\sum_{m=1}^{M} \mathbf{p}_{t}^{(m)} \\ 
            &= \mathbf{w}_{t-1} - \frac{1}{M}\sum_{m=1}^{M} \eta F\left(\mathbf{w}_{t-\frac{1}{2}}^{(m)};\xi_{t}^{(m)}\right) - \frac{1}{M}\sum_{m=1}^{M} \mathbf{e}_{t-1}^{(m)} \\ 
            &= \tilde{\mathbf{w}}_{t-1} - \frac{1}{M}\sum_{m=1}^{M} \eta F\left(\mathbf{w}_{t-\frac{1}{2}}^{(m)};\xi_{t}^{(m)}\right) . 
        \end{aligned}
    \end{equation}
    Suppose that $\mathbf{w}^*$ is an optimum solution. 
    \begin{equation}
        \label{eq:main1}
        \begin{aligned}
            &\| \tilde{\mathbf{w}}_{t} - \mathbf{w}^* \|^2 
            = \| \tilde{\mathbf{w}}_{t-1} - \frac{1}{M}\sum_{m=1}^{M} \eta F\left(\mathbf{w}_{t-\frac{1}{2}}^{(m)};\xi_{t}^{(m)}\right) - \mathbf{w}^* \|^2 \\ 
            &= \| \tilde{\mathbf{w}}_{t-1} - \frac{1}{M}\sum_{m=1}^{M} \eta F\left(\mathbf{w}_{t-\frac{1}{2}}^{(m)};\xi_{t}^{(m)}\right) - \mathbf{w}^* \|^2 \\
            & \quad\quad - \| \tilde{\mathbf{w}}_{t-1} - \frac{1}{M}\sum_{m=1}^{M} \eta F\left(\mathbf{w}_{t-\frac{1}{2}}^{(m)};\xi_{t}^{(m)}\right) - \tilde{\mathbf{w}}_{t} \|^2 \\ 
            &= \| \tilde{\mathbf{w}}_{t-1} - \mathbf{w}^* \|^2 - \| \tilde{\mathbf{w}}_{t-1} - \tilde{\mathbf{w}}_{t} \|^2 
             + 2 \left\langle \mathbf{w}^* - \tilde{\mathbf{w}}_{t}, \frac{1}{M}\sum_{m=1}^{M} \eta F\left(\mathbf{w}_{t-\frac{1}{2}}^{(m)};\xi_{t}^{(m)}\right) \right\rangle \\ 
            &= \| \tilde{\mathbf{w}}_{t-1} - \mathbf{w}^* \|^2 - \| \tilde{\mathbf{w}}_{t-1} - \tilde{\mathbf{w}}_{t} \|^2 
             + 2 \left\langle \mathbf{w}^* - \bar{\mathbf{w}}_{t-\frac{1}{2}}, \frac{1}{M}\sum_{m=1}^{M} \eta F\left(\mathbf{w}_{t-\frac{1}{2}}^{(m)};\xi_{t}^{(m)}\right) \right\rangle \\
            & \quad\quad + 2 \left\langle \bar{\mathbf{w}}_{t-\frac{1}{2}} - \tilde{\mathbf{w}}_{t}, \frac{1}{M}\sum_{m=1}^{M} \eta F\left(\mathbf{w}_{t-\frac{1}{2}}^{(m)};\xi_{t}^{(m)}\right) \right\rangle \\ 
            &= \| \tilde{\mathbf{w}}_{t-1} - \mathbf{w}^* \|^2 - \| \tilde{\mathbf{w}}_{t-1} - \bar{\mathbf{w}}_{t-\frac{1}{2}} + \bar{\mathbf{w}}_{t-\frac{1}{2}} - \tilde{\mathbf{w}}_{t} \|^2 \\ 
            & \quad\quad + 2 \left\langle \mathbf{w}^* - \bar{\mathbf{w}}_{t-\frac{1}{2}}, \frac{1}{M}\sum_{m=1}^{M} \eta F\left(\mathbf{w}_{t-\frac{1}{2}}^{(m)};\xi_{t}^{(m)}\right) \right\rangle \\
            & \quad\quad + 2 \left\langle \bar{\mathbf{w}}_{t-\frac{1}{2}} - \tilde{\mathbf{w}}_{t}, \frac{1}{M}\sum_{m=1}^{M} \eta F\left(\mathbf{w}_{t-\frac{1}{2}}^{(m)};\xi_{t}^{(m)}\right) \right\rangle \\ 
            &= \| \tilde{\mathbf{w}}_{t-1} - \mathbf{w}^* \|^2 - \| \tilde{\mathbf{w}}_{t-1} - \bar{\mathbf{w}}_{t-\frac{1}{2}} \|^2 - \| \bar{\mathbf{w}}_{t-\frac{1}{2}} - \tilde{\mathbf{w}}_{t} \|^2 \\ 
            & \quad\quad + 2 \left\langle \mathbf{w}^* - \bar{\mathbf{w}}_{t-\frac{1}{2}}, \frac{1}{M}\sum_{m=1}^{M} \eta F\left(\mathbf{w}_{t-\frac{1}{2}}^{(m)};\xi_{t}^{(m)}\right) \right\rangle \\
            & \quad\quad + 2 \left\langle \tilde{\mathbf{w}}_{t} - \bar{\mathbf{w}}_{t-\frac{1}{2}}, \tilde{\mathbf{w}}_{t-1} - \bar{\mathbf{w}}_{t-\frac{1}{2}} - \frac{1}{M}\sum_{m=1}^{M} \eta F\left(\mathbf{w}_{t-\frac{1}{2}}^{(m)};\xi_{t}^{(m)}\right) \right\rangle \\
        \end{aligned}
    \end{equation}
    \\
    Note that
    \begin{equation}
        \label{eq:pesudo}
        \begin{aligned}
            &\mathbb{E} \left[ 2 \left\langle \mathbf{w}^* - \bar{\mathbf{w}}_{t-\frac{1}{2}}, \frac{1}{M}\sum_{m=1}^{M} \eta F\left(\mathbf{w}_{t-\frac{1}{2}}^{(m)};\xi_{t}^{(m)}\right) \right\rangle \right]\\
            &= \mathbb{E} \left[ 2 \left\langle \mathbf{w}^* - \bar{\mathbf{w}}_{t-\frac{1}{2}}, \frac{1}{M}\sum_{m=1}^{M} \eta F\left(\mathbf{w}_{t-\frac{1}{2}}^{(m)}\right) \right\rangle \right]\\
            & \quad\quad + \mathbb{E} \left[ 2 \left\langle \mathbf{w}^* - \bar{\mathbf{w}}_{t-\frac{1}{2}}, \frac{1}{M}\sum_{m=1}^{M} \eta F\left(\mathbf{w}_{t-\frac{1}{2}}^{(m)};\xi_{t}^{(m)}\right) - \frac{1}{M}\sum_{m=1}^{M} \eta F\left(\mathbf{w}_{t-\frac{1}{2}}^{(m)}\right) \right\rangle \right]\\
            &\overset{(a)}{\leq} \mathbb{E} \left[ 2 \left\langle \mathbf{w}^* - \bar{\mathbf{w}}_{t-\frac{1}{2}}, \frac{1}{M}\sum_{m=1}^{M} \eta F\left(\mathbf{w}_{t-\frac{1}{2}}^{(m)};\xi_{t}^{(m)}\right) - \frac{1}{M}\sum_{m=1}^{M} \eta F\left(\mathbf{w}_{t-\frac{1}{2}}^{(m)}\right) \right\rangle \right]
            \overset{(b)}{=} 0
        \end{aligned}
    \end{equation}
    where $(a)$ holds by the pseudomonotonicity of operator $F$, $(b)$ hold since $\mathbb{E} \left[ F(\mathbf{w}_{t-\frac{1}{2}}^{(m)};\xi_{t,b}^{(m)}) \right] = F(\mathbf{w}_{t-\frac{1}{2}}^{(m)})$ ($1\leq b \leq B$). 
    Since \ref{eq:tilde}, we have
    \begin{equation}
        \label{eq:111}
        \begin{aligned}
            &2 \left\langle \tilde{\mathbf{w}}_{t} - \bar{\mathbf{w}}_{t-\frac{1}{2}}, \tilde{\mathbf{w}}_{t-1} - \bar{\mathbf{w}}_{t-\frac{1}{2}} - \frac{1}{M}\sum_{m=1}^{M} \eta F\left(\mathbf{w}_{t-\frac{1}{2}}^{(m)};\xi_{t}^{(m)}\right) \right\rangle\\
            &= 2 \left\langle \tilde{\mathbf{w}}_{t} - \bar{\mathbf{w}}_{t-\frac{1}{2}}, 
            \tilde{\mathbf{w}}_{t} - \bar{\mathbf{w}}_{t-\frac{1}{2}} \right\rangle 
            = 2 \| \tilde{\mathbf{w}}_{t} - \bar{\mathbf{w}}_{t-\frac{1}{2}} \|^{2} \\
            &= 2 \| \tilde{\mathbf{w}}_{t-1} - \bar{\mathbf{w}}_{t-\frac{1}{2}} - \frac{1}{M}\sum_{m=1}^{M} \eta F\left(\mathbf{w}_{t-\frac{1}{2}}^{(m)};\xi_{t}^{(m)}\right) \|^2 \\
        \end{aligned}
    \end{equation}
    In addition, according to the update rules in Algorithm~\ref{alg:DQOSG}, we have
    \begin{equation}
        \label{eq:222}
        \bar{\mathbf{w}}_{t-\frac{1}{2}} = \mathbf{w}_{t-1} - \frac{1}{M}\sum_{m=1}^{M} \eta F\left(\mathbf{w}_{t-\frac{3}{2}}^{(m)};\xi_{t-1}^{(m)}\right) - \frac{1}{M}\sum_{m=1}^{M} \mathbf{e}_{t-1}^{(m)}
    \end{equation}
    We combine~\ref{eq:111} and~\ref{eq:222}, which yield
    \begin{equation}
        \label{eq:updaterule}
        \begin{aligned}
            &2 \left\langle \tilde{\mathbf{w}}_{t} - \bar{\mathbf{w}}_{t-\frac{1}{2}}, \tilde{\mathbf{w}}_{t-1} - \bar{\mathbf{w}}_{t-\frac{1}{2}} - \frac{1}{M}\sum_{m=1}^{M} \eta F\left(\mathbf{w}_{t-\frac{1}{2}}^{(m)};\xi_{t}^{(m)}\right) \right\rangle\\
            &= 2 \| \tilde{\mathbf{w}}_{t-1} - \left[ \tilde{\mathbf{w}}_{t-1} - \frac{1}{M}\sum_{m=1}^{M} \eta F\left(\mathbf{w}_{t-\frac{3}{2}}^{(m)};\xi_{t-1}^{(m)}\right) \right] - \frac{1}{M}\sum_{m=1}^{M} \eta F\left(\mathbf{w}_{t-\frac{1}{2}}^{(m)};\xi_{t}^{(m)}\right) \|^2 \\
            &= 2 \| \frac{1}{M}\sum_{m=1}^{M} \eta F\left(\mathbf{w}_{t-\frac{3}{2}}^{(m)};\xi_{t-1}^{(m)}\right) - \frac{1}{M}\sum_{m=1}^{M} \eta F\left(\mathbf{w}_{t-\frac{1}{2}}^{(m)};\xi_{t}^{(m)}\right) \|^2 \\
            &\overset{(a)}{\leq} 6 \eta^2 \| \frac{1}{M}\sum_{m=1}^{M} F\left(\mathbf{w}_{t-\frac{3}{2}}^{(m)}\right) - \frac{1}{M}\sum_{m=1}^{M} F\left(\mathbf{w}_{t-\frac{1}{2}}^{(m)}\right) \|^2 
             + 6 \eta^2 \| \bar{\mathbf{\epsilon}}_{t} \|^2 + 6 \eta^2 \| \bar{\mathbf{\epsilon}}_{t-1} \|^2 \\
            &\overset{(b)}{\leq} 18 \eta^2 \| \frac{1}{M}\sum_{m=1}^{M} F\left(\mathbf{w}_{t-\frac{3}{2}}^{(m)}\right) - F\left(\bar{\mathbf{w}}_{t-\frac{3}{2}}\right) \|^2 
             + 18 \eta^2 \| F\left(\bar{\mathbf{w}}_{t-\frac{3}{2}}\right) - F\left(\bar{\mathbf{w}}_{t-\frac{1}{2}}\right) \|^2 \\
            & \quad\quad\quad\quad + 18 \eta^2 \| F\left(\bar{\mathbf{w}}_{t-\frac{1}{2}}\right) - \frac{1}{M}\sum_{m=1}^{M} F\left(\mathbf{w}_{t-\frac{1}{2}}^{(m)}\right) \|^2 
             + 6 \eta^2 \| \bar{\mathbf{\epsilon}}_{t} \|^2 + 6 \eta^2 \| \bar{\mathbf{\epsilon}}_{t-1} \|^2 \\
            &\overset{(c)}{\leq} 18 \eta^2 \| \frac{1}{M}\sum_{m=1}^{M} F\left(\mathbf{w}_{t-\frac{3}{2}}^{(m)}\right) - F\left(\bar{\mathbf{w}}_{t-\frac{3}{2}}\right) \|^2 
             + 18 \eta^2 L^2 \| \bar{\mathbf{w}}_{t-\frac{3}{2}} - \bar{\mathbf{w}}_{t-\frac{1}{2}} \|^2 \\
            & \quad\quad\quad\quad + 18 \eta^2 \| F\left(\bar{\mathbf{w}}_{t-\frac{1}{2}}\right) - \frac{1}{M}\sum_{m=1}^{M} F\left(\mathbf{w}_{t-\frac{1}{2}}^{(m)}\right) \|^2 
             + 6 \eta^2 \| \bar{\mathbf{\epsilon}}_{t} \|^2 + 6 \eta^2 \| \bar{\mathbf{\epsilon}}_{t-1} \|^2 \\
            &\overset{(d)}{\leq} 18 \eta^2 \| \frac{1}{M}\sum_{m=1}^{M} F\left(\mathbf{w}_{t-\frac{3}{2}}^{(m)}\right) - F\left(\bar{\mathbf{w}}_{t-\frac{3}{2}}\right) \|^2 \\
            & \quad\quad\quad\quad + 18 \eta^2 \| F\left(\bar{\mathbf{w}}_{t-\frac{1}{2}}\right) - \frac{1}{M}\sum_{m=1}^{M} F\left(\mathbf{w}_{t-\frac{1}{2}}^{(m)}\right) \|^2 \\
            & \quad\quad\quad\quad + 36 \eta^2 L^2 \left( \| \bar{\mathbf{w}}_{t-\frac{3}{2}} - \tilde{\mathbf{w}}_{t-1} \|^2 + \| \tilde{\mathbf{w}}_{t-1} - \bar{\mathbf{w}}_{t-\frac{1}{2}} \|^2 \right) \\
            & \quad\quad\quad\quad + 6 \eta^2 \| \bar{\mathbf{\epsilon}}_{t} \|^2 + 6 \eta^2 \| \bar{\mathbf{\epsilon}}_{t-1} \|^2 \\
        \end{aligned}
    \end{equation}
    where 
    $(a)$ and $(b)$ hold since $ \|\mathbf{a}+\mathbf{b}+\mathbf{c}\|^2 \leq 3\|\mathbf{a}\|^2 + 3\|\mathbf{b}\|^2 + 3\|\mathbf{c}\|^2 $,
    $(c)$ holds by the $L$-Lipschitz continuity of $F$, 
    $(d)$ holds since $\|\mathbf{a}+\mathbf{b}\|^2 \leq 2\|\mathbf{a}\|^2 + 2\|\mathbf{b}\|^2$.
    \\
    \\
    We combine (\ref{eq:main1}), (\ref{eq:pesudo}) and (\ref{eq:updaterule}), which yield
    \begin{equation}
        \label{eq:main2}
        \begin{aligned}
            &\mathbb{E} \left[ \| \tilde{\mathbf{w}}_{t} - \mathbf{w}^* \|^2 \right] \\
            &\leq \mathbb{E} \left[ \| \tilde{\mathbf{w}}_{t-1} - \mathbf{w}^* \|^2 \right] - \mathbb{E} \left[ \| \tilde{\mathbf{w}}_{t-1} - \bar{\mathbf{w}}_{t-\frac{1}{2}} \|^2 \right] - \mathbb{E} \left[ \| \bar{\mathbf{w}}_{t-\frac{1}{2}} - \tilde{\mathbf{w}}_{t} \|^2 \right] \\ 
            & \quad + 18 \eta^2 \mathbb{E} \left[ \| \frac{1}{M}\sum_{m=1}^{M} F\left(\mathbf{w}_{t-\frac{3}{2}}^{(m)}\right) - F\left(\bar{\mathbf{w}}_{t-\frac{3}{2}}\right) \|^2 \right] \\
            & \quad + 18 \eta^2 \mathbb{E} \left[ \| F\left(\bar{\mathbf{w}}_{t-\frac{1}{2}}\right) - \frac{1}{M}\sum_{m=1}^{M} F\left(\mathbf{w}_{t-\frac{1}{2}}^{(m)}\right) \|^2 \right] \\
            & \quad  + 36 \eta^2 L^2 \left( \mathbb{E} \left[ \| \bar{\mathbf{w}}_{t-\frac{3}{2}} - \tilde{\mathbf{w}}_{t-1} \|^2 \right] + \mathbb{E} \left[ \| \tilde{\mathbf{w}}_{t-1} - \bar{\mathbf{w}}_{t-\frac{1}{2}} \|^2 \right] \right) \\
            & \quad + 6 \eta^2 \mathbb{E} \left[ \| \bar{\mathbf{\epsilon}}_{t} \|^2 \right] + 6 \eta^2 \mathbb{E} \left[ \| \bar{\mathbf{\epsilon}}_{t-1} \|^2 \right] \\
        \end{aligned}
    \end{equation}
    Note that 
    \begin{equation}
        \label{eq:lemma-average}
        \begin{aligned}
            &\mathbb{E} \left[ \| \frac{1}{M}\sum_{m=1}^{M} F\left(\mathbf{w}_{t-\frac{1}{2}}^{(m)}\right) - F\left(\bar{\mathbf{w}}_{t-\frac{1}{2}}\right)\|^{2} \right] 
            \leq \frac{1}{M}\sum_{m=1}^{M} \mathbb{E} \left[ \| F\left(\mathbf{w}_{t-\frac{1}{2}}^{(m)}\right) - F\left(\bar{\mathbf{w}}_{t-\frac{1}{2}}\right)\|^{2} \right] \\
            &\overset{(a)}{\leq} \frac{L^2}{M}\sum_{m=1}^{M} \mathbb{E} \left[ \| \mathbf{w}_{t-\frac{1}{2}}^{(m)} - \bar{\mathbf{w}}_{t-\frac{1}{2}}\|^{2} \right] \\
            &\overset{(b)}{=} \frac{L^2}{M}\sum_{m=1}^{M} \mathbb{E} [ \| \left( \mathbf{w}_{t-1} - \eta F\left(\mathbf{w}_{t-\frac{3}{2}}^{(m)};\xi_{t-1}^{(m)}\right) - \mathbf{e}_{t-1}^{(m)} \right) - ( \mathbf{w}_{t-1} - \\
            & \quad\quad\quad\quad \frac{1}{M}\sum_{m^{'}=1}^{M} \eta F\left(\mathbf{w}_{t-\frac{3}{2}}^{(m^{'})};\xi_{t-1}^{(m^{'})}\right) - \frac{1}{M}\sum_{m^{'}=1}^{M} \mathbf{e}_{t-1}^{(m^{'})} ) ~\|^{2} ] \\
            &= \frac{L^2}{M}\sum_{m=1}^{M} \mathbb{E} [ \| \left( \frac{1}{M}\sum_{m^{'}=1}^{M} \eta F\left(\mathbf{w}_{t-\frac{3}{2}}^{(m^{'})};\xi_{t-1}^{(m^{'})}\right) - \eta F\left(\mathbf{w}_{t-\frac{3}{2}}^{(m)};\xi_{t-1}^{(m)}\right) \right) \\
            & \quad\quad\quad\quad + \left( \frac{1}{M}\sum_{m^{'}=1}^{M} \mathbf{e}_{t-1}^{(m^{'})} - \mathbf{e}_{t-1}^{(m)} \right) \|^{2} ] \\
            &\leq \frac{2 L^2 \eta^2}{M}\sum_{m=1}^{M} \mathbb{E} \left[ \| \frac{1}{M}\sum_{m^{'}=1}^{M} F\left(\mathbf{w}_{t-\frac{3}{2}}^{(m^{'})};\xi_{t-1}^{(m^{'})}\right) - F\left(\mathbf{w}_{t-\frac{3}{2}}^{(m)};\xi_{t-1}^{(m)}\right) \|^2 \right] \\
            & \quad\quad\quad\quad + \frac{2 L^2}{M}\sum_{m=1}^{M} \mathbb{E} \left[ \| \frac{1}{M}\sum_{m^{'}=1}^{M} \mathbf{e}_{t-1}^{(m^{'})} - \mathbf{e}_{t-1}^{(m)} \|^{2} \right] \\
            &\overset{(c)}{\leq} \frac{6 L^2 \eta^2}{M}\sum_{m=1}^{M} \mathbb{E} \left[ \| \frac{1}{M}\sum_{m^{'}=1}^{M} F\left(\mathbf{w}_{t-\frac{3}{2}}^{(m^{'})};\xi_{t-1}^{(m^{'})}\right) - \frac{1}{M}\sum_{m^{'}=1}^{M} F\left(\mathbf{w}_{t-\frac{3}{2}}^{(m^{'})}\right) \|^2 \right] \\
            & \quad\quad\quad\quad + \frac{6 L^2 \eta^2}{M}\sum_{m=1}^{M} \mathbb{E} \left[ \| \frac{1}{M}\sum_{m^{'}=1}^{M} F\left(\mathbf{w}_{t-\frac{3}{2}}^{(m^{'})}\right) - F\left(\mathbf{w}_{t-\frac{3}{2}}^{(m)}\right) \|^2 \right] \\
            & \quad\quad\quad\quad + \frac{6 L^2 \eta^2}{M}\sum_{m=1}^{M} \mathbb{E} \left[  \| F\left(\mathbf{w}_{t-\frac{3}{2}}^{(m)}\right) - F\left(\mathbf{w}_{t-\frac{3}{2}}^{(m)};\xi_{t-1}^{(m)}\right) \|^2 \right] \\
            & \quad\quad\quad\quad + \frac{2 L^2}{M}\sum_{m=1}^{M} \mathbb{E} \left[ \| \frac{1}{M}\sum_{m^{'}=1}^{M} \mathbf{e}_{t-1}^{(m^{'})} - \mathbf{e}_{t-1}^{(m)} \|^{2} \right] \\
            &= 6 L^2 \eta^2 \mathbb{E} \left[ \| \bar{\mathbf{\epsilon}}_{t-1} \|^2 \right] 
             + \frac{6 L^2 \eta^2}{M}\sum_{m=1}^{M} \mathbb{E} \left[ \| \frac{1}{M}\sum_{m^{'}=1}^{M} F\left(\mathbf{w}_{t-\frac{3}{2}}^{(m^{'})}\right) - F\left(\mathbf{w}_{t-\frac{3}{2}}^{(m)}\right) \|^2 \right] \\
            & \quad\quad\quad\quad + \frac{6 L^2 \eta^2}{M}\sum_{m=1}^{M} \mathbb{E} \left[ \| \mathbf{\epsilon}_{t-1}^{(m)} \|^2 \right] 
             + \frac{2 L^2}{M}\sum_{m=1}^{M} \mathbb{E} \left[ \| \frac{1}{M}\sum_{m^{'}=1}^{M} \mathbf{e}_{t-1}^{(m^{'})} - \mathbf{e}_{t-1}^{(m)} \|^{2} \right] \\
            &\leq \frac{12 L^2 \eta^2 \sigma^2}{BM} 
             + \frac{24 L^2 \eta^2 G^2 (M-1)}{M} 
             + \frac{64 L^2 \eta^2(1-\delta)(G^{2}+\frac{\sigma^2}{B})(M-1)}{\delta^2 M} \\
            &\overset{(d)}{\leq}
             \frac{12 L^2 \sigma^2}{B^2M^2} 
             + \frac{24 L^2 G^2 (M-1)}{BM^2} 
             + \frac{64 L^2 (1-\delta)(G^{2}+\frac{\sigma^2}{B})(M-1)}{\delta^2 BM^2}
        \end{aligned}
    \end{equation}
    where $(a)$ holds by the $L$-Lipschitz continuity of $F$, 
    $(b)$ holds by the update of the algorithm, 
    $(c)$ holds since $ \|\mathbf{a}+\mathbf{b}+\mathbf{c}\|^2 \leq 3\|\mathbf{a}\|^2 + 3\|\mathbf{b}\|^2 + 3\|\mathbf{c}\|^2 $, 
    $(d)$ holds since $\eta \leq \min \{ \frac{1}{\sqrt{BM}}, \frac{1}{6\sqrt{2}L} \}$.
    By employing (\ref{eq:main2}) and (\ref{eq:lemma-average}), we have
    \begin{equation}
        \label{eq:main3}
        \begin{aligned}
            &\mathbb{E} \left[ \| \tilde{\mathbf{w}}_{t} - \mathbf{w}^* \|^2 \right] \\
            &\leq \mathbb{E} \left[ \| \tilde{\mathbf{w}}_{t-1} - \mathbf{w}^* \|^2 \right] - \mathbb{E} \left[ \| \tilde{\mathbf{w}}_{t-1} - \bar{\mathbf{w}}_{t-\frac{1}{2}} \|^2 \right] - \mathbb{E} \left[ \| \bar{\mathbf{w}}_{t-\frac{1}{2}} - \tilde{\mathbf{w}}_{t} \|^2 \right] \\
            & \quad + \left( \frac{432 L^2 \sigma^2}{B^2M^2} + \frac{864 L^2 G^2 (M-1)}{BM^2} 
             + \frac{2304 L^2 (1-\delta)(G^{2}+\frac{\sigma^2}{B})(M-1)}{\delta^2 BM^2} \right) \eta^2 \\
            & \quad + 36 \eta^2 L^2 \mathbb{E} \left[ \| \bar{\mathbf{w}}_{t-\frac{3}{2}} - \tilde{\mathbf{w}}_{t-1} \|^{2} + \| \tilde{\mathbf{w}}_{t-1} - \bar{\mathbf{w}}_{t-\frac{1}{2}} \|^{2}\right] 
             + \frac{12 \eta^2 \sigma^2}{BM}\\
        \end{aligned}
    \end{equation}
    We rearrange terms in (\ref{eq:main3}), which yield
    \begin{equation}
        \begin{aligned}
            &\mathbb{E} \left[ \| \tilde{\mathbf{w}}_{t-1} - \bar{\mathbf{w}}_{t-\frac{1}{2}} \|^2 \right] 
            + \mathbb{E} \left[ \| \bar{\mathbf{w}}_{t-\frac{1}{2}} - \tilde{\mathbf{w}}_{t} \|^2 \right] \\
            & \quad - 36 \eta^2 L^2 \mathbb{E} \left[ \| \bar{\mathbf{w}}_{t-\frac{3}{2}} - \tilde{\mathbf{w}}_{t-1} \|^{2} + \| \tilde{\mathbf{w}}_{t-1} - \bar{\mathbf{w}}_{t-\frac{1}{2}} \|^{2} \right] \\
            &\leq \mathbb{E} \left[ \| \tilde{\mathbf{w}}_{t-1} - \mathbf{w}^* \|^2 - \| \tilde{\mathbf{w}}_{t} - \mathbf{w}^* \|^2 \right] + \frac{12\eta^2\sigma^2}{BM} \\
            & \quad + \left( \frac{432 L^2 \sigma^2}{B^2M^2} + \frac{864 L^2 G^2 (M-1)}{BM^2} 
             + \frac{2304 L^2 (1-\delta)(G^{2}+\frac{\sigma^2}{B})(M-1)}{\delta^2 BM^2} \right) \eta^2 \\
        \end{aligned}
    \end{equation}
    Then we have
    \begin{equation}
        \begin{aligned}
            & \frac{1}{T} \sum_{t=1}^{T} \left( 1 - 36\eta^2L^2\right) 
            \mathbb{E} \left[  \| \tilde{\mathbf{w}}_{t-1} - \bar{\mathbf{w}}_{t-\frac{1}{2}} \|^2 + \| \bar{\mathbf{w}}_{t-\frac{1}{2}} - \tilde{\mathbf{w}}_{t} \|^2 \right] \leq \frac{\| \tilde{\mathbf{w}}_{0} - \mathbf{w}^* \|^2}{T} \\
            & \quad + \left( \frac{432 L^2 \sigma^2}{B^2M^2} + \frac{864 L^2 G^2 (M-1)}{BM^2} 
             + \frac{2304 L^2 (1-\delta)(G^{2}+\frac{\sigma^2}{B})(M-1)}{\delta^2 BM^2} + \frac{12\sigma^2}{BM} \right) \eta^2 
        \end{aligned}
    \end{equation}
    Since $\eta \leq \min\{\frac{1}{\sqrt{BM}},\frac{1}{6\sqrt{2}L}\}$, we have $1 - 36 L^{2} \eta^{2} \geq \frac{1}{2}$. Then we have 
    \begin{equation}
        \label{eq:main4}
        \begin{aligned}
            &\frac{1}{2T} \sum_{t=1}^{T} 
            \mathbb{E} \left[  \| \tilde{\mathbf{w}}_{t-1} - \bar{\mathbf{w}}_{t-\frac{1}{2}} \|^2 + \| \bar{\mathbf{w}}_{t-\frac{1}{2}} - \tilde{\mathbf{w}}_{t} \|^2 \right] \leq \frac{\| \tilde{\mathbf{w}}_{0} - \mathbf{w}^* \|^2}{T} \\
            & \quad + \left( \frac{432 L^2 \sigma^2}{B^2M^2} + \frac{864 L^2 G^2 (M-1)}{BM^2} 
             + \frac{2304 L^2 (1-\delta)(G^{2}+\frac{\sigma^2}{B})(M-1)}{\delta^2 B M^2} + \frac{12 \sigma^2}{BM} \right) \eta^2 
        \end{aligned}
    \end{equation}
    Note that
    \begin{equation}
        \label{eq:error-corrected}
        \begin{aligned}
            \frac{1}{T} \sum_{t=1}^{T} \mathbb{E} & \left[ \eta^{2} \| \frac{1}{M}\sum_{m=1}^{M} F\left(\mathbf{w}_{t-\frac{1}{2}}^{(m)};\xi_{t}^{(m)}\right) \|^2 \right] \\
            &= \frac{1}{T} \sum_{t=1}^{T} \mathbb{E} \left[ \| \tilde{\mathbf{w}}_{t-1} - \tilde{\mathbf{w}}_{t} \|^2 \right] \\
            &\leq \frac{1}{T} \sum_{t=1}^{T} \mathbb{E} \left[ 2 \| \tilde{\mathbf{w}}_{t-1} - \bar{\mathbf{w}}_{t-\frac{1}{2}} \|^2 + 2 \| \bar{\mathbf{w}}_{t-\frac{1}{2}} - \tilde{\mathbf{w}}_{t} \|^2 \right] \\
        \end{aligned}
    \end{equation}
    By employing (\ref{eq:main4}) and (\ref{eq:error-corrected}),
    we have
    \begin{equation}
        \begin{aligned}
            &\frac{1}{T} \sum_{t=1}^{T} \mathbb{E} \left[ \| \frac{1}{M}\sum_{m=1}^{M} F\left(\mathbf{w}_{t-\frac{1}{2}}^{(m)};\xi_{t}^{(m)}\right) \|^2 \right] 
            \leq \frac{4 \| \tilde{\mathbf{w}}_{0} - \mathbf{w}^* \|^2}{\eta^{2} T} 
            + \frac{1728~ L^2 \sigma^2}{B^2M^2} \\
            & \quad\quad\quad\quad + \frac{3456~ L^2 G^2 (M-1)}{BM^2} + \frac{9216~ L^2 (1-\delta)(G^{2}+\frac{\sigma^2}{B})(M-1)}{\delta^2 BM^2} + \frac{48~ \sigma^2}{BM}
        \end{aligned}
    \end{equation}
	
\end{proof}
\end{document}